\documentclass[a4paper,conference]{IEEEtran}

\usepackage{amsmath}
\usepackage{amssymb}
\usepackage{graphicx}
\usepackage{cite}
\usepackage[pagebackref,breaklinks,colorlinks]{hyperref}

\begin{document}

\title{Scene Aware Person Image Generation through Global Contextual Conditioning}

\author{\IEEEauthorblockN{Prasun Roy\IEEEauthorrefmark{1},
Subhankar Ghosh\IEEEauthorrefmark{1},
Saumik Bhattacharya\IEEEauthorrefmark{2},
Umapada Pal\IEEEauthorrefmark{3}, and
Michael Blumenstein\IEEEauthorrefmark{1}}
\IEEEauthorblockA{\IEEEauthorrefmark{1}University of Technology Sydney, Australia}
\IEEEauthorblockA{\IEEEauthorrefmark{2}Indian Institute of Technology Kharagpur, India}
\IEEEauthorblockA{\IEEEauthorrefmark{3}Indian Statistical Institute Kolkata, India}
\IEEEauthorblockA{\url{https://prasunroy.github.io}}}

\maketitle

\begin{abstract}
Person image generation is an intriguing yet challenging problem. However, this task becomes even more difficult under constrained situations. In this work, we propose a novel pipeline to generate and insert contextually relevant person images into an existing scene while preserving the global semantics. More specifically, we aim to insert a person such that the location, pose, and scale of the person being inserted blends in with the existing persons in the scene. Our method uses three individual networks in a sequential pipeline. At first, we predict the potential location and the skeletal structure of the new person by conditioning a Wasserstein Generative Adversarial Network (WGAN) on the existing human skeletons present in the scene. Next, the predicted skeleton is refined through a shallow linear network to achieve higher structural accuracy in the generated image. Finally, the target image is generated from the refined skeleton using another generative network conditioned on a given image of the target person. In our experiments, we achieve high-resolution photo-realistic generation results while preserving the general context of the scene. We conclude our paper with multiple qualitative and quantitative benchmarks on the results.
\end{abstract}

\section{Introduction}\label{sec:introduction}

Person image generation is an essential step of many computer vision applications such as pose and motion transfer, virtual try-on, etc. However, the problem is challenging as the generation algorithms have to render the output images from limited contextual information. Researchers have made dedicated efforts to design algorithms that can generate realistic images of a person with different unobserved poses from a single reference image of that person \cite{ma2017pose,ma2018disentangled,siarohin2018deformable,esser2018variational,balakrishnan2018synthesizing,zhao2018multi,zhu2019progressive}. Although these algorithms can generate fairly realistic outputs, they mainly focus on a single human body for transferring poses. In our work, we try to push the boundaries of the person generation problem by making it aware of the global context information. We aim to insert a generated person image into a given scene where multiple human figures are already present. Consequently, the proposed algorithm needs to consider multiple contextual information simultaneously, such as spatial location, pose, scale, and occlusion. This makes the generation task more challenging yet more flexible for real-world applications.

To address the overall complexity of the task, we divide the problem into three individual sub-problems. At first, we assume the global context as the geometric information expressed as a set of skeletal structures corresponding to the human figures present in the scene. Each human skeleton is a set of spatial coordinates of 18 body keypoints. Therefore, we represent the global context by encoding the existing human skeletal structures as an 18-channel many-hot heatmap where each channel corresponds to one specific body keypoint. We approximate the spatial location and pose of the target person by conditioning a WGAN model over the global context encoding. Next, we use a fully-connected linear network to refine the small spatial deviations in the crude target pose estimated in the previous step. This simple refinement step helps us significantly improve visual quality and photo-realism in the final generated image. To accommodate unknown distortions, we train the refinement network in a semi-supervised manner. Finally, the target person image is generated from the refined target pose obtained in the second stage, and the resulting image is inserted into the scene following the positional information received in the first stage. We adopt a multi-scale attention guided pose transfer network \cite{roy2022multi} to retain both coarse and fine details in the generated image.

The major contributions of our work are as follows:

\begin{itemize}
    \item We propose a novel person image generation technique that can insert a person into a given scene while maintaining the general contextual relevancy in the resulting image.
    \item The proposed method can handle the presence of different numbers of people in the given scene. The generation pipeline is adaptable to spatial locations, pose variations, and scales and can generate context-preserving high-resolution photo-realistic images.
\end{itemize}

To the best of our knowledge, this work is one of the first attempts to design a pipeline for person image generation that tries to preserve the global contextual information of the source.

The remainder of the paper is organized as follows. Sec. \ref{sec:related_work} introduces the state-of-the-art person image generation techniques in different application domains. In Sec. \ref{sec:methodology}, we discuss the proposed pipeline in detail along with the individual stages. Sec. \ref{sec:dataset_and_training} presents our experimental settings during training and evaluation of the proposed pipeline. We also introduce the dataset used to train and evaluate the models. In Sec. \ref{sec:results}, we present the qualitative and quantitative analysis of the generation results. In Sec. \ref{sec:limitations}, we report some limitations of the proposed pipeline. Finally, in Sec. \ref{sec:conclusion}, we conclude the paper by summarizing our findings and potential prospects.

\section{Related Work}\label{sec:related_work}

In computer vision, novel view synthesis is a widely explored problem domain. The introduction of Generative Adversarial Networks (GANs) \cite{goodfellow2014generative} has been highly influential for the recent progress in perceptually realistic image synthesis \cite{goodfellow2014generative,mirza2014conditional,johnson2016perceptual,radford2016unsupervised,lassner2017generative,ledig2017photo}. Conditional GANs \cite{mirza2014conditional,isola2017image,sangkloy2017scribbler,zhu2017unpaired} have been adopted in several application domains in computer vision, such as inpainting \cite{yeh2017semantic}, super-resolution \cite{dong2015image, kim2016accurate} etc. Person image generation and human pose transfer are conditional generative problems, where the target image is generated from a target pose by conditioning the generative process on a given source image. With the progress in conditional generative modeling, the performance of such algorithms has improved significantly in recent years. The initial approaches of person image generation \cite{ma2017pose,ma2018disentangled} have introduced disentangled multi-stage pipelines. In \cite{zhao2018multi}, the authors propose a method for generating multi-view images of a person from a single observation in a coarse to fine approach. In \cite{balakrishnan2018synthesizing}, the authors introduce a pose transfer scheme to segment and generate the foreground and background separately. In another technique \cite{wang2018toward}, a geometric matching module is used in a characteristic preserving generative network. The coarse to fine generation technique is further improved by disentangling the foreground, background, and pose into separate branches of a generative network \cite{ma2018disentangled}. In \cite{pumarola2018unsupervised}, the authors propose an unsupervised multi-level generation method with a pose conditioned bidirectional generator. In \cite{zanfir2018human}, the authors first estimate a 3D mesh from a single image followed by pose transfer using the mesh. Another algorithm \cite{siarohin2018deformable} introduces deformable GANs with a nearest-neighbor loss for transferring pose. In \cite{esser2018variational}, authors propose a variational U-Net for conditional appearance and shape generation. Researchers have also explored 3D approximations, and manifold learning \cite{neverova2018dense,li2019dense} to transfer the pose. More recently, an attention-guided progressive generation scheme \cite{zhu2019progressive} has been introduced by leveraging the attention mechanism to transfer pose progressively in an end-to-end manner. In \cite{li2020pona}, the authors propose pose-guided non-local attention with a long-range dependency to progressively select important regions of the image. Gafni \textit{et al.} \cite{gafni2020wish} propose the only existing pipeline for context-aware person image synthesis by utilizing a mask-based 3-stage generation framework.

\section{Methodology}\label{sec:methodology}

\begin{figure*}[ht]
  \centering
  \includegraphics[width=\linewidth]{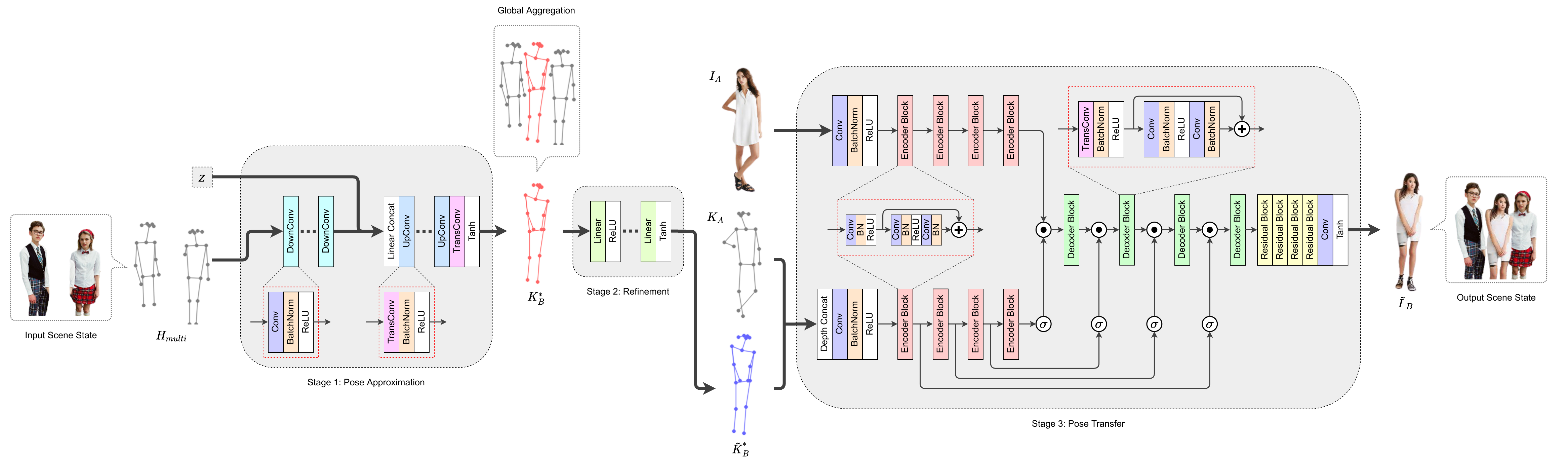}
  \caption{Architecture of the proposed pipeline. The workflow is disentangled among three sequential stages. In stage 1, an approximation of the target pose is estimated by sampling from a Gaussian distribution conditioned over the global geometric context. Next, the crude representation is refined by regression in stage 2. Finally, the pose transfer is carried out by conditioning over the source image in stage 3.}
  \label{fig:architecture}
\end{figure*}

The pipeline for the proposed technique is divided among three independent sub-networks. In the first stage, we estimate an approximation for the spatial location and pose of the target. To achieve a contextually relevant representation, we derive the target geometry by sampling from a Gaussian distribution and conditioning it over the global geometric context of the existing actors in the scene. This crude target pose is then refined through regression in the next stage. Finally, we perform pose transfer by conditioning the transformation on an existing image of the target actor. In Fig. \ref{fig:architecture}, we show an overview of our integrated generation pipeline.

\subsection{Context Preserving Pose Approximation}

Assuming a given scene that contains images of $n$ human actors, the goal of our algorithm is to introduce another actor into the scene such that the new actor \emph{blends in} with the existing $n$ actors by preserving the global scene context. We represent the global geometric context of such a scene by a set $P_n$ of $n$ spatial pose representations corresponding to all $n$ existing actors. We represent every individual pose $P_i$ as another set of $k$ body-joint locations (keypoints). For a scene image of dimension $w \times h$, we represent every individual pose $P_i$ as an $w \times h \times k$ binary heatmap $H_i$, where each channel of the heatmap corresponds to one specific keypoint. The spatial location of every visible keypoint is encoded as 1 in the respective position of $H_i$ with 0 everywhere else (\emph{one-hot}). We represent the global geometric scene context as an aggregate of all such $n$ heatmaps. This results in a binary heatmap $H_n$ of dimension $w \times h \times k$ where each channel contains spatial encoding for a specific keypoint of all $n$ existing actors (\emph{many-hot}). We use an isotropic representation for each channel of $H_n$ by estimating the Gaussian of the Euclidean distance from the keypoint to reduce the high sparsity of $H_n$. We denote such representation of the global context as $H_{multi}$.

We use a WGAN model to estimate the potential location and the pose of the target actor by sampling from a normal distribution conditioned on the global geometric scene context $H_{multi}$. Like conventional GAN models, we have a generator network $G_T$ and a discriminator network $D_T$. In $G_T$, we first encode $H_{multi}$ to a 512-dimension vector $v_B$ by down-scaling through consecutive convolution layers. To allow variations in the generated poses, we sample a noise vector $z \sim \textit{N}(\mathbf{0}, I)$, where $I$ is the identity matrix of size $512 \times 512$, and we use it as another input to $G_T$. Both $v_B$ and $z$ are linearly concatenated and passed through 4 up-convolution layers. Each of the up-convolution layers contain a transposed convolution followed by batch normalization \cite{ioffe2015batch} and ReLU activation \cite{nair2010rectified}. The transposed convolution layers in the upscaling branch contain 256, 128, 64, and 32 filters, respectively. The output of the final up-convolution layer is passed through another transposed convolution layer followed by $tanh$ activation to produce the final output of $G_T$. The final transposed convolution layer contains 18 filters such that we obtain a generated output $G_T(v_B, z)$ of dimension $64 \times 64 \times 18$, where each channel represents one of the 18 keypoints $k_j$, $j \in \{1, 2, \hdots, 18\}$. As the training is performed adversarially, we also employ a discriminator $D_T$ that evaluates the quality of the generated keypoint set. In $D_T$, we pass $G_T(v_B, z)$ through 4 consecutive convolution layers where each layer has a stride of 2 and is followed by leaky ReLU activation. We have 32, 64, 128, and 256 filters in these convolution layers, respectively. Finally, the scalar output of the discriminator is estimated using another convolution layer with a single channel and stride 4.

The optimization objective $L_D$ for the discriminator $D_T$ is given by
\begin{equation*}
  L_D = - \mathbb{E}_{(x, v_B) \sim p_t, z \sim p_z} [D_T(x, v_B) - D_T(G_T(z, v_B), v_B)]
\end{equation*}
where $(x, v_B) \sim p_t$ is the heatmap and the corresponding global context encoding pair sampled from the training set, $z \sim p_z$ is the noise vector sampled from a Gaussian distribution.

For Wasserstein Generative Adversarial Networks, researchers \cite{gulrajani2017improved} have demonstrated that the discriminator objective should be Lipschitz continuous to achieve a stable training. In this case, the discriminator acts more like a critic to reflect the \emph{realness} of the generated samples. To enforce the Lipschitz constraint, we compute the Gradient Penalty \cite{gulrajani2017improved} as
\begin{equation*}
  \mathcal{P}_T = \mathbb{E}_{(\tilde{x}, v_B) \sim p_{\tilde{x}, v_B}} [(\|\nabla_{\tilde{x}, v_B} D_T(\tilde{x}, v_B)\|_2-1)^2]
\end{equation*}
where $\|.\|_2$ indicates the $l_2$ norm and $\tilde{x}$ is a synthetic sample obtained from the weighted sum of a real sample $x$ and a generated sample $G_T(z, v_B)$, mathematically, $\tilde{x} = \alpha G_T(z, v_B) + (1 - \alpha)x$, where $\alpha$ is a random number, selected from a uniform distribution between 0 and 1. As indicated by \cite{gulrajani2017improved}, $\mathcal{P}_T$ tries to restrict the gradient magnitude to 1 and helps to enforce the Lipschitz constraint. After including the Gradient Penalty term, the final loss function for $D_T$ becomes
\begin{equation*}
  L_{D_T} = L_D + \lambda \mathcal{P}_T
\end{equation*}
where $\lambda$ is the regularizing constant that controls the effect of gradient penalty $\mathcal{P}_T$ on the overall discriminator loss $L_{D_T}$. In all our experiments, we set $\lambda = 10$.

To optimize the generator network $G_T$, we define the loss function $L_{G_T}$ as
\begin{equation*}
  L_{G_T} = -\mathbb{E}_{v_B \sim p_{v_B}, z \sim p_z} [D_T(G_T(z, v_B), v_B)]
\end{equation*}

The generator $G_T$ estimates the probable location and pose of the target person as an 18-channel heatmap. We estimate the precise spatial locations of individual keypoints from this heatmap. To represent the target pose as a set of keypoints $K^*_B$, we compute the maximum activation $\psi^{max}_j$, $j \in \{1, 2, \hdots, 18\}$ for every channel. The position of the maximum activation for $j$-th channel is assumed to be the spatial location of $j$-th keypoint if $\psi^{max}_j \geq 0.2$. The $j$-th keypoint is considered occluded if $\psi^{max}_j < 0.2$.

\subsection{Pose Refinement}

The generator $G_T$ estimates an initial approximation of the target pose represented as the set of keypoints $K^*_B$. Due to potential uncertainty in the position and location of the target person, the keypoints $K^*_B$ can have spatial perturbations. We have noticed that such spatial noise is most prominent for the facial keypoints (nose, two eyes, and two ears). As we have adopted a keypoint-guided pose transfer scheme in the next stage, even small fluctuations in the coordinates of the facial keypoints result in a visually unrealistic generation. To mitigate the effects of such perturbations, we use a linear fully-connected network $\omega_R$ (RefineNet) for refining the facial keypoints by regression. First, we translate the five facial keypoints $k^f_i$, $i \in \{1, 2, \hdots, 5\}$ by $(k^f_i - k_n)$ where $k_n$ is the spatial location of the nose. Next, we normalize the translated facial keypoints such that the scaled keypoints $k^s_i$ are within a cuboid of span $\pm 1$ with the scaled nose keypoint at the origin (0, 0). Then, we flatten the scaled facial keypoints to a 10-dimensional vector $v_f$ and pass it through three fully-connected linear hidden layers, each having 128 nodes followed by ReLU activation. We have 10 nodes at the output layer followed by $tanh$ activation. The network is optimized by minimizing the mean squared error (MSE) between the actual and the predicted keypoints. While training, we augment $k^s_i$ with small random 2D spatial perturbations and try to predict the original values of $k^s_i$. Finally, we denormalize and retranslate the predicted facial keypoints. The refined set of keypoints $\tilde{K}^*_B$ is obtained by updating the coordinates of the facial keypoints with the predictions from RefineNet.

\subsection{Multi-scale Attention Guided Pose Transfer}

We use a conditional GAN consisting of a generator network $G_R$ and a discriminator network $D_R$ for transferring pose. We estimate the source pose as the set of keypoints $K_A$ extracted from the source image $I_A$ using an existing human pose estimator \cite{cao2017realtime}. We then construct two sparse heatmaps $H_A$ and $\tilde{H}_B$ corresponding to $K_A$ and $\tilde{K}^*_B$, respectively. The generator $G_R$ takes the condition image $I_A$ of dimension $256 \times 256 \times 3$ and the channel-wise concatenated pose heatmaps $(H_A, \tilde{H}_B)$ of dimension $256 \times 256 \times 36$ as inputs to generate an output image $\tilde{I}_B$ of dimension $256 \times 256 \times 3$ as an estimate for the target image $I_B$. We use a Markovian PatchGAN discriminator \cite{isola2017image} as $D_R$ to evaluate the visual correctness of the generated image. The discriminator takes two channel-wise concatenated images, either $(I_A, I_B)$ or $(I_A, \tilde{I}_B)$, of dimension $256 \times 256 \times 6$ as input and estimates a binary class probability map for $70 \times 70$ input patches.

In $G_R$, we pass $I_A$ and $(H_A, \tilde{H}_B)$ through two separate encoding branches. At first, the input is projected to a $256 \times 256 \time 64$ feature space by convolution (kernel size = $3 \times 3$, stride = 1, padding = 1, bias = 0), batch normalization, and ReLU activation in each branch. Then, we downscale the resulting feature maps through 4 consecutive encoding blocks. We denote the encoder blocks as $\theta^i_l$ for image branch and $\theta^h_l$ for pose heatmaps branch, where $l \in \{1, 2, \hdots, 4\}$. At each subsequent encoding block, we downscale the feature space dimension to half and double the number of filters. Each encoding block consists of convolution (kernel size = $4 \times 4$, stride = 2, padding = 1, bias = 0), batch normalization, and ReLU activation, followed by a basic residual block \cite{he2016deep}. We merge the output feature maps from the encoding branches and upscale the combined feature space by passing through a single decoding branch consisting of 4 consecutive blocks. We denote the decoder blocks as $\phi_l$, where $l \in \{1, 2, \hdots, 4\}$. We upscale the feature space dimension by two at each subsequent decoding block and reduce the number of filters by half. Each decoding block consists of transposed convolution (kernel size = $4 \times 4$, stride = 2, padding = 1, bias = 0), batch normalization, and ReLU activation, followed by a basic residual block \cite{he2016deep}. We construct dense attention links between downstream and upstream branches at every resolution level to preserve coarse and fine details in the generated image. Mathematically, at the lowest resolution level, where $l = 4$,
\begin{equation*}
  I^\phi_3 = \phi_4 (I^{\theta^i}_4 \;\odot\; \sigma(H^{\theta^h}_4))
\end{equation*}
and for the higher resolution levels, where $l = \{1, 2, 3\}$,
\begin{equation*}
  I^\phi_{l-1} = \phi_l (I^{\theta^i}_l \;\odot\; \sigma(H^{\theta^h}_l))
\end{equation*}
where $I^\phi_l$ is the output from decoder block $\phi_l$, $I^{\theta^i}_l$ is the output from image encoder block $\theta^i_l$, $H^{\theta^h}_l$ is the output from pose encoder block $\theta^h_l$, $\sigma$ denotes an element-wise \emph{sigmoid} activation function, and $\odot$ denotes an element-wise product. Finally, the resulting feature maps are passed through 4 consecutive residual blocks followed by a point-wise convolution (kernel size = $1 \times 1$, stride = 1, padding = 0, bias = 0) to project the feature space into an output of dimension $256 \times 256 \times 3$. We apply the hyperbolic tangent activation function \emph{tanh} on the output tensor to get the normalized generated image $\tilde{I}_B$.

We update the parameters of $G_R$ by optimizing three objectives -- pixel-wise $l_1$ loss $L^{G_R}_{l_1}$, discriminator loss $L^{G_R}_{GAN}$ estimated from $D_R$, and perceptual loss $L^{G_R}_{P_\rho}$ computed using a pre-trained VGG-19 network \cite{simonyan2015very}. We define the pixel-wise $l_1$ loss as $L^{G_R}_{l_1} = \|\tilde{I}_B - I_B\|_1$, where $\|.\|_1$ is the $l_1$ norm. The discriminator loss is defined as
\begin{equation*}
  L^{G_R}_{GAN} = \mathcal{L}_{BCE}(D_R(I_A, \tilde{I}_B), 1)
\end{equation*}
where $\mathcal{L}_{BCE}$ is the binary cross-entropy loss. Finally, we compute the perceptual loss as
\begin{equation*}
  L^{G_R}_{P_\rho} = \frac{1}{h_\rho w_\rho c_\rho} \sum_{x=1}^{h_\rho} \sum_{y=1}^{w_\rho} \sum_{z=1}^{c_\rho} \|q_\rho(\tilde{I}_B) - q_\rho(I_B)\|_1
\end{equation*}
where $q_\rho$ is the output of the $\rho$-th layer of the VGG-19 network having a feature space of dimension $(h_\rho \times w_\rho \times c_\rho)$. In our approach, we include two perceptual loss terms for $\rho = 4$ and $\rho = 9$ into the overall objective function. So, the optimization objective for $G_R$ is given by
\begin{equation*}
  L_{G_R} = \lambda_1 L^{G_R}_{l_1} + \lambda_2 L^{G_R}_{GAN} + \lambda_3 (L^{G_R}_{P_4} + L^{G_R}_{P_9})
\end{equation*}
where $\lambda_1$, $\lambda_2$ and $\lambda_3$ are the weighing constants for respective loss terms. In our experiments, we have taken $\lambda_1 = 5$, $\lambda_2 = 1$, and $\lambda_3 = 5$. Lastly, we define the optimization objective of $D_R$ as
\begin{equation*}
  L_{D_R} = \frac{1}{2} \left[\mathcal{L}_{BCE}(D_R(I_A, I_B), 1) + \mathcal{L}_{BCE}(D_R(I_A,\tilde{I}_B), 0)\right]
\end{equation*}

\section{Dataset and Training}\label{sec:dataset_and_training}

We use a multi-human parsing dataset MHP-v1 \cite{li2017multiple} to train the pose approximation model. The dataset contains 4980 images where multiple persons appear in contextually correlated poses for each image. We first use a pre-trained human pose estimator \cite{cao2017realtime} to derive the keypoints of individual persons present in each image. While training, one random person is selected as the target, and the remaining persons are used as source actors. To train both pose refinement and pose transfer networks, we use the DeepFashion \cite{liu2016deepfashion} dataset. Out of 40488 images, 37344 images are used for training and 3144 images for testing.

We optimize both generator and discriminator of the WGAN (stage-1) using the stochastic Adam optimizer \cite{kingma2015adam} with learning rate $\eta_1 = 1e^{-4}$, $\beta_1 = 0$, $\beta_2 = 0.9$, $\epsilon = 1e^{-8}$, and weight decay = 0. While training, we update the generator once for every five updates of the discriminator to prevent mode collapse. The linear pose refinement network (stage-2) is optimized using stochastic gradient descent with learning rate $\eta_2 = 1e^{-2}$. The multi-scale attention-guided pose transfer network (stage-3) is also optimized with the stochastic Adam optimizer \cite{kingma2015adam}, but with learning rate $\eta_3 = 1e^{-3}$, $\beta_1 = 0.5$, $\beta_2 = 0.999$, $\epsilon = 1e^{-8}$, and weight decay = 0. We initialize the parameters of every generator and discriminator before training by sampling from a normal distribution of 0 mean and 0.02 standard deviation.

\section{Results}\label{sec:results}

\begin{figure*}[ht]
  \centering
  \includegraphics[width=0.90\linewidth]{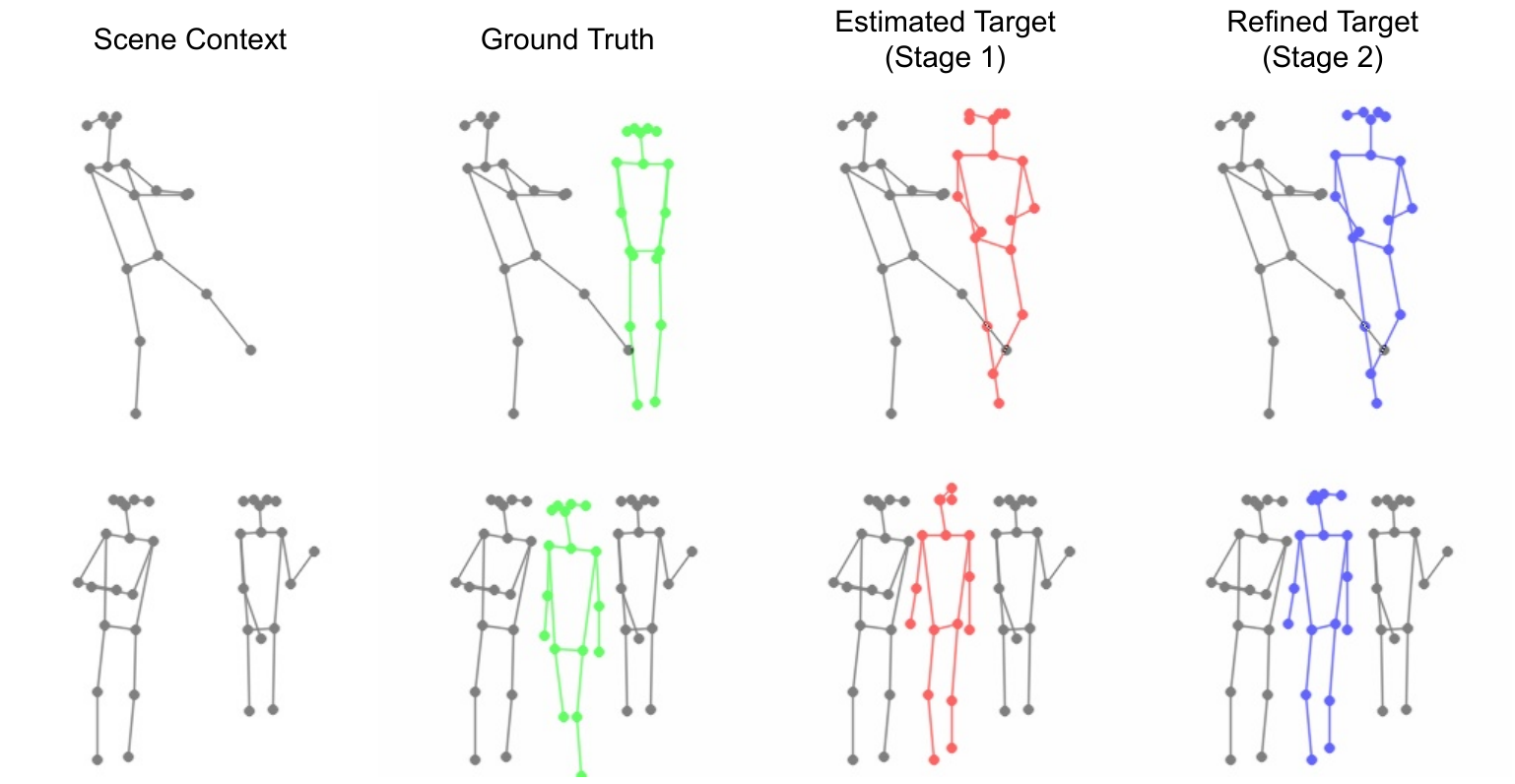}
  \caption{Qualitative results of pose approximation followed by pose refinement. Due to spatial location and pose uncertainty, the target pose may look different from the ground truth. However, it does not affect the generation performance as long as the global geometric context is preserved and the target person blends in with the existing persons in the scene.}
  \label{fig:result_1}
\end{figure*}

\begin{figure}[ht]
  \centering
  \includegraphics[width=\linewidth]{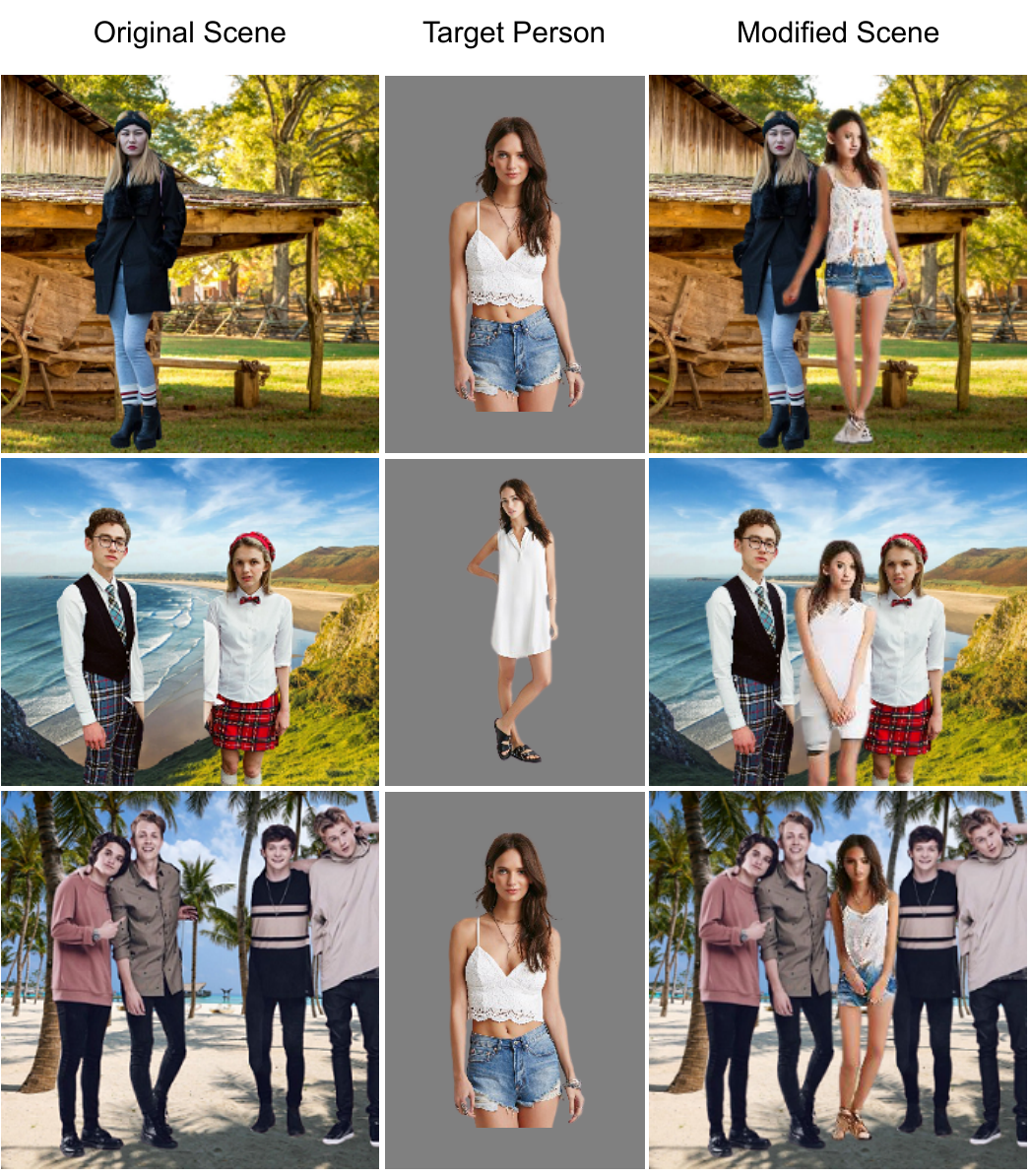}
  \caption{Qualitative results generated by the proposed pipeline.}
  \label{fig:result_2}
\end{figure}

\begin{figure}[ht]
  \centering
  \includegraphics[width=\linewidth]{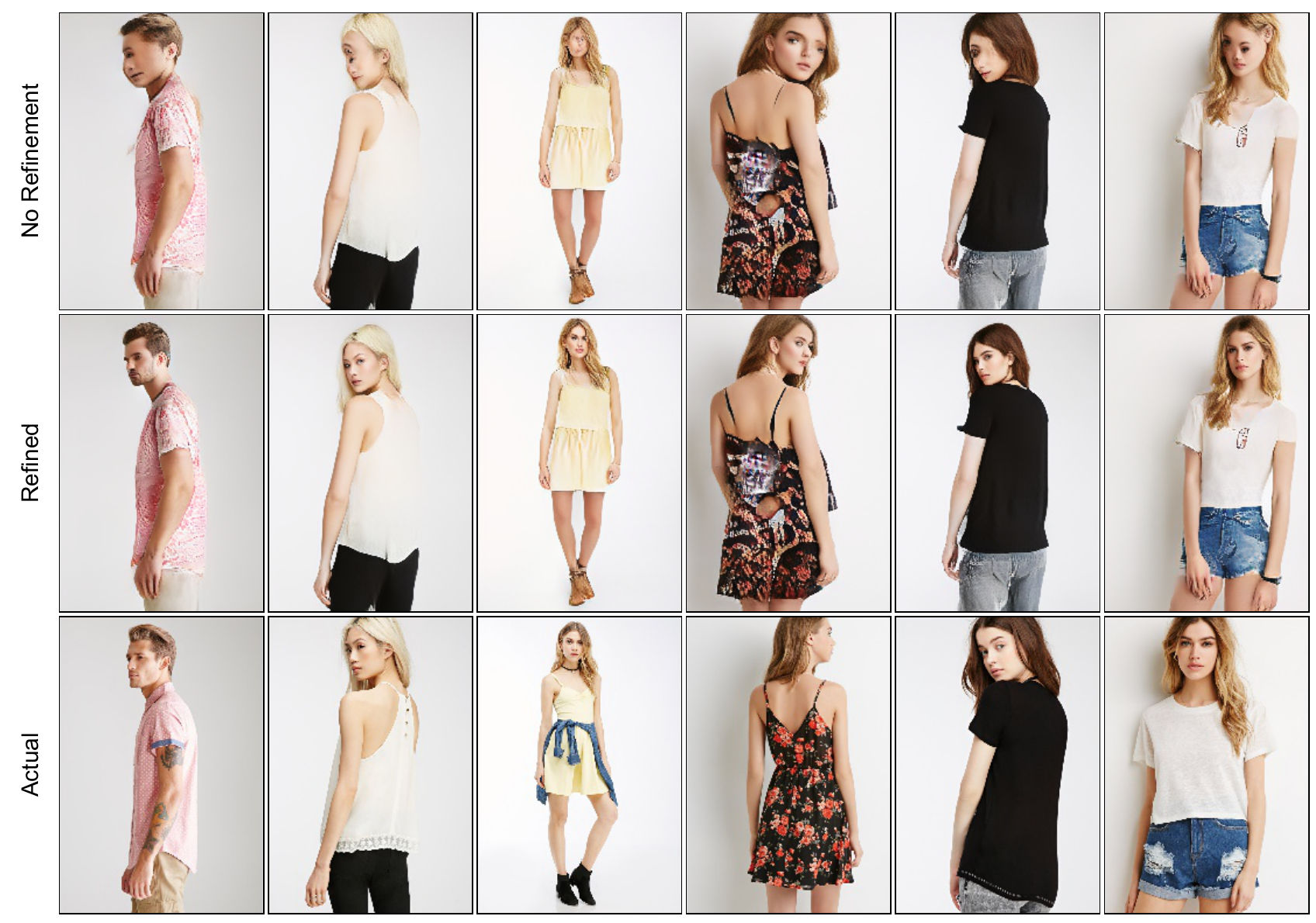}
  \caption{Effects of intermediate pose refinement on generated images.}
  \label{fig:result_3}
\end{figure}

In Fig. \ref{fig:result_1}, we show the qualitative results of pose approximation and pose refinement. Our method can estimate the spatial location and pose of the target person realistically while keeping the global contextual consistency with the existing persons in the scene. An important point to note is that there is uncertainty in the geometry of the target person as the target can adopt a wide range of locations and pose orientations. This natural uncertainty is introduced by sampling the target pose from a Gaussian distribution. Therefore, the predicted target geometry may be significantly different from the ground truth, but this does not affect the generation performance as long as the target blends in with the existing persons in the scene, preserving the overall scene context.

In Fig. \ref{fig:result_2}, we show the qualitative results of the entire pipeline consisting of all three stages. The proposed method can synthesize high quality contextually relevant images of the target person that blends in well with the existing persons present in the scene.

In Fig. \ref{fig:result_3}, we show the necessity of the pose refinement performed in stage 2. The quality of the generated image in stage 3 is heavily dependent upon the spatial adjustments performed during this refinement. It significantly reduces facial distortions and thereby produces more photo-realistic results.

At present, a quantifiable generalized metric for the visual quality assessment of images is an open problem in computer vision. However, in the previously proposed pose transfer algorithms \cite{ma2017pose,siarohin2018deformable,esser2018variational,zhu2019progressive}, the authors have estimated a few widely used evaluation metrics for quantifying visual quality. This includes the Structural Similarity Index Measure (SSIM) \cite{wang2004image}, Inception Score (IS) \cite{salimans2016improved}, Detection Score (DS) \cite{liu2016ssd} and PCKh \cite{andriluka20142d}. SSIM measures the perceived quality of the generated images by comparing them with the respective real images and considering image degradation as a perceived change in the structural information. IS uses Inception architecture \cite{szegedy2015going} as an image classifier to estimate the KL divergence \cite{kullback1951information} between the label distribution and the marginal distribution for a large set of images. DS uses an object detector to estimate the target class recognition confidence of the object detection model as a measure of the perceptual quality. PCKh aims to quantify the shape consistency between the generated and real person images by estimating the percentage of correctly aligned keypoints. We also evaluate the Learned Perceptual Image Patch Similarity (LPIPS) \cite{zhang2018unreasonable} metric, which is a more modern standard for assessing perceptual image quality.

In Table \ref{tab:comparison}, we show an analytical comparison among different pose transfer methods by evaluating SSIM, IS, DS, PCKh, and LPIPS scores. For SSIM and IS, we obtain slightly lower scores than the best results. However, for DS, PCKh, and LPIPS, our method has achieved the best scores. We improve the PCKh score over PATN \cite{zhu2019progressive} by 2\% indicating a superior shape consistency with better alignment among keypoints. Similar to \cite{gafni2020wish}, we also evaluate LPIPS score for both VGG16 \cite{simonyan2015very} and SqueezeNet \cite{iandola2016squeezenet} backbones. Our method achieves significantly better scores in both cases, indicating an improved perceptual quality over other methods.

\begin{table}[t]
\caption{Quantitative comparison among different pose transfer methods -- $\text{PG}^2$ \cite{ma2017pose}, Deformable GANs \cite{siarohin2018deformable}, VUNet \cite{esser2018variational}, PATN \cite{zhu2019progressive} and our method.}
\centering
\resizebox{\columnwidth}{!}{%
\begin{tabular}{l|cccccc}
\hline
Method & SSIM $\uparrow$ & IS $\uparrow$ & DS $\uparrow$ & PCKh $\uparrow$ & \begin{tabular}[c]{@{}c@{}}LPIPS $\downarrow$\\ (VGG)\end{tabular} & \begin{tabular}[c]{@{}c@{}}LPIPS $\downarrow$\\ (SqzNet)\end{tabular} \\ \hline
$\text{PG}^2$ \cite{ma2017pose}      & 0.773 & 3.163 & 0.951 & 0.89 & 0.523 & 0.416 \\
Deform \cite{siarohin2018deformable} & 0.760 & 3.362 & 0.967 & 0.94 & - & - \\
VUNet \cite{esser2018variational}    & 0.763 & \textbf{3.440} & 0.972 & 0.93 & - & - \\
PATN \cite{zhu2019progressive}       & \textbf{0.773} & 3.209 & 0.976 & 0.96 & 0.299 & 0.170 \\
Ours                                 & 0.769 & 3.379 & \textbf{0.976} & \textbf{0.98} & \textbf{0.200} & \textbf{0.111} \\ \hline
Real Data                            & 1.000 & 3.864 & 0.974 & 1.00 & 0.000 & 0.000 \\ \hline
\end{tabular}%
}
\label{tab:comparison}
\end{table}

\section{Limitations}\label{sec:limitations}

\begin{figure}[ht]
  \centering
  \includegraphics[width=\linewidth]{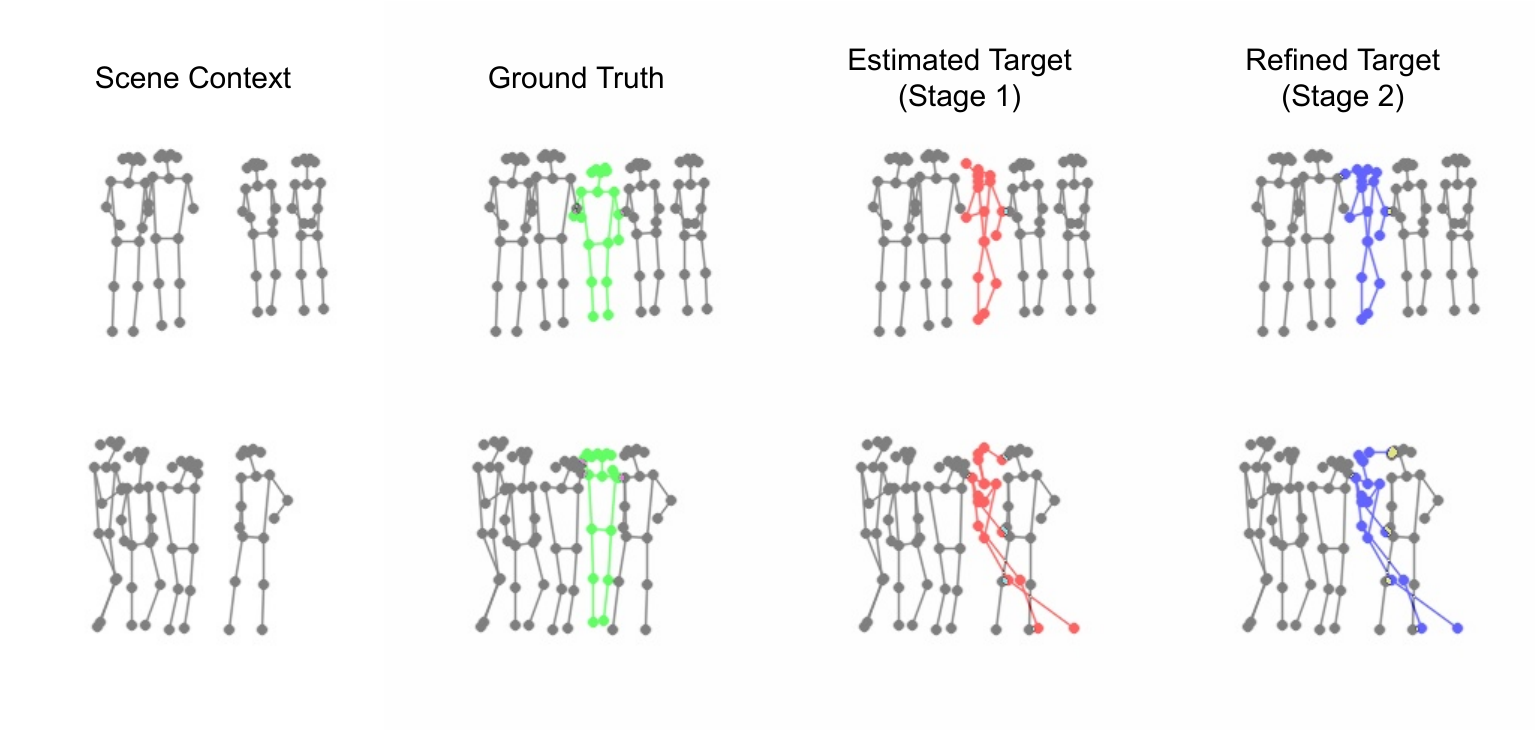}
  \caption{Examples of failure cases.}
  \label{fig:limitations}
\end{figure}

On some occasions, the proposed method fails to perform well. In particular, for crowded scenes with many persons, the pipeline struggles to generate photo-realistic results. In these situations, the WGAN generator fails to estimate an appropriate spatial location and a realistic pose for the target person in stage 1. This problem subsequently causes an unrealistic target image generation during pose transfer in stage 3. Therefore, the generative capability of the pose transfer stage is heavily dependent on the success of pose approximation and pose refinement stages in the proposed pipeline. We show a few such failure cases in Fig. \ref{fig:limitations}.

\section{Conclusion}\label{sec:conclusion}

In this work, we have proposed a three-stage sequential pipeline for generating and inserting a person image into an existing scene. Our algorithm can estimate semantically relevant spatial position and pose for the target person. Apart from preserving the global scene context, our method also produces high-quality photo-realistic human images. To the best of our knowledge, the proposed method is one of the first attempts to generate scene-aware person images that can maintain the synergy between the rendered person and the existing human figures in the scene. To handle the overall complexity of the method, we have divided the generation architecture into three individual components. First, we estimate the potential spatial position and the pose of the target person by conditioning a WGAN on the global scene context. Next, we regressively refine the initial approximation of the target pose with a linear fully-connected network. Finally, we generate the target image from the refined target pose by using a given image of the target person as a conditioning criterion for the GAN. We construct the resulting blended scene by inserting the target person image generated in stage 3 at the spatial location obtained in stage 1. Also, we have conducted the relevant visual and analytical comparisons to justify our method. Lastly, we conclude our paper with a discussion on the limitations of the proposed method. The generative capability of our approach is restricted by the spatial location and pose approximation phase. In our experiments, we have found that the spatial approximation struggles for an overpopulated scene which ultimately leads to a poor target image generation. We plan to explore a way to circumvent this problem in the future.

\section*{Acknowledgements}

This work is partially supported by the Technology Innovation Hub, Indian Statistical Institute, Kolkata, India.

\bibliographystyle{IEEEtran}
\bibliography{references}

\begin{thebibliography}{10}
\providecommand{\url}[1]{#1}
\csname url@samestyle\endcsname
\providecommand{\newblock}{\relax}
\providecommand{\bibinfo}[2]{#2}
\providecommand{\BIBentrySTDinterwordspacing}{\spaceskip=0pt\relax}
\providecommand{\BIBentryALTinterwordstretchfactor}{4}
\providecommand{\BIBentryALTinterwordspacing}{\spaceskip=\fontdimen2\font plus
\BIBentryALTinterwordstretchfactor\fontdimen3\font minus \fontdimen4\font\relax}
\providecommand{\BIBforeignlanguage}[2]{{%
\expandafter\ifx\csname l@#1\endcsname\relax
\typeout{** WARNING: IEEEtran.bst: No hyphenation pattern has been}%
\typeout{** loaded for the language `#1'. Using the pattern for}%
\typeout{** the default language instead.}%
\else
\language=\csname l@#1\endcsname
\fi
#2}}
\providecommand{\BIBdecl}{\relax}
\BIBdecl

\bibitem{ma2017pose}
L.~Ma, X.~Jia, Q.~Sun, B.~Schiele, T.~Tuytelaars, and L.~Van~Gool, ``Pose guided person image generation,'' in \emph{The Conference on Neural Information Processing Systems (NIPS)}, 2017.

\bibitem{ma2018disentangled}
L.~Ma, Q.~Sun, S.~Georgoulis, L.~Van~Gool, B.~Schiele, and M.~Fritz, ``Disentangled person image generation,'' in \emph{The IEEE Conference on Computer Vision and Pattern Recognition (CVPR)}, 2018.

\bibitem{siarohin2018deformable}
A.~Siarohin, E.~Sangineto, S.~Lathuili{\`e}re, and N.~Sebe, ``Deformable {GAN}s for pose-based human image generation,'' in \emph{The IEEE Conference on Computer Vision and Pattern Recognition (CVPR)}, 2018.

\bibitem{esser2018variational}
P.~Esser, E.~Sutter, and B.~Ommer, ``A variational u-net for conditional appearance and shape generation,'' in \emph{The IEEE Conference on Computer Vision and Pattern Recognition (CVPR)}, 2018.

\bibitem{balakrishnan2018synthesizing}
G.~Balakrishnan, A.~Zhao, A.~V. Dalca, F.~Durand, and J.~Guttag, ``Synthesizing images of humans in unseen poses,'' in \emph{The IEEE Conference on Computer Vision and Pattern Recognition (CVPR)}, 2018.

\bibitem{zhao2018multi}
B.~Zhao, X.~Wu, Z.-Q. Cheng, H.~Liu, Z.~Jie, and J.~Feng, ``Multi-view image generation from a single-view,'' in \emph{The ACM International Conference on Multimedia (MM)}, 2018.

\bibitem{zhu2019progressive}
Z.~Zhu, T.~Huang, B.~Shi, M.~Yu, B.~Wang, and X.~Bai, ``Progressive pose attention transfer for person image generation,'' in \emph{The IEEE Conference on Computer Vision and Pattern Recognition (CVPR)}, 2019.

\bibitem{roy2022multi}
P.~Roy, S.~Bhattacharya, S.~Ghosh, and U.~Pal, ``Multi-scale attention guided pose transfer,'' \emph{arXiv preprint arXiv:2202.06777}, 2022.

\bibitem{goodfellow2014generative}
I.~Goodfellow, J.~Pouget-Abadie, M.~Mirza, B.~Xu, D.~Warde-Farley, S.~Ozair, A.~Courville, and Y.~Bengio, ``Generative adversarial nets,'' in \emph{The Conference on Neural Information Processing Systems (NIPS)}, 2014.

\bibitem{mirza2014conditional}
M.~Mirza and S.~Osindero, ``Conditional generative adversarial nets,'' \emph{arXiv preprint arXiv:1411.1784}, 2014.

\bibitem{johnson2016perceptual}
J.~Johnson, A.~Alahi, and L.~Fei-Fei, ``Perceptual losses for real-time style transfer and super-resolution,'' in \emph{The European Conference on Computer Vision (ECCV)}, 2016.

\bibitem{radford2016unsupervised}
A.~Radford, L.~Metz, and S.~Chintala, ``Unsupervised representation learning with deep convolutional generative adversarial networks,'' in \emph{The International Conference on Learning Representations (ICLR)}, 2016.

\bibitem{lassner2017generative}
C.~Lassner, G.~Pons-Moll, and P.~V. Gehler, ``A generative model of people in clothing,'' in \emph{The IEEE International Conference on Computer Vision (ICCV)}, 2017.

\bibitem{ledig2017photo}
C.~Ledig, L.~Theis, F.~Husz{\'a}r, J.~Caballero, A.~Cunningham, A.~Acosta, A.~Aitken, A.~Tejani, J.~Totz, Z.~Wang, and W.~Shi, ``Photo-realistic single image super-resolution using a generative adversarial network,'' in \emph{The IEEE Conference on Computer Vision and Pattern Recognition (CVPR)}, 2017.

\bibitem{isola2017image}
P.~Isola, J.-Y. Zhu, T.~Zhou, and A.~A. Efros, ``{I}mage-to-{I}mage translation with conditional adversarial networks,'' in \emph{The IEEE Conference on Computer Vision and Pattern Recognition (CVPR)}, 2017.

\bibitem{sangkloy2017scribbler}
P.~Sangkloy, J.~Lu, C.~Fang, F.~Yu, and J.~Hays, ``Scribbler: Controlling deep image synthesis with sketch and color,'' in \emph{The IEEE Conference on Computer Vision and Pattern Recognition (CVPR)}, 2017.

\bibitem{zhu2017unpaired}
J.-Y. Zhu, T.~Park, P.~Isola, and A.~A. Efros, ``Unpaired image-to-image translation using cycle-consistent adversarial networks,'' in \emph{The IEEE International Conference on Computer Vision (ICCV)}, 2017.

\bibitem{yeh2017semantic}
R.~A. Yeh, C.~Chen, T.~Yian~Lim, A.~G. Schwing, M.~Hasegawa-Johnson, and M.~N. Do, ``Semantic image inpainting with deep generative models,'' in \emph{The IEEE Conference on Computer Vision and Pattern Recognition (CVPR)}, 2017.

\bibitem{dong2015image}
C.~Dong, C.~C. Loy, K.~He, and X.~Tang, ``Image super-resolution using deep convolutional networks,'' \emph{IEEE Transactions on Pattern Analysis and Machine Intelligence (TPAMI)}, 2015.

\bibitem{kim2016accurate}
J.~Kim, J.~Kwon~Lee, and K.~Mu~Lee, ``Accurate image super-resolution using very deep convolutional networks,'' in \emph{The IEEE Conference on Computer Vision and Pattern Recognition (CVPR)}, 2016.

\bibitem{wang2018toward}
B.~Wang, H.~Zheng, X.~Liang, Y.~Chen, L.~Lin, and M.~Yang, ``Toward characteristic-preserving image-based virtual try-on network,'' in \emph{The European Conference on Computer Vision (ECCV)}, 2018.

\bibitem{pumarola2018unsupervised}
A.~Pumarola, A.~Agudo, A.~Sanfeliu, and F.~Moreno-Noguer, ``Unsupervised person image synthesis in arbitrary poses,'' in \emph{The IEEE Conference on Computer Vision and Pattern Recognition (CVPR)}, 2018.

\bibitem{zanfir2018human}
M.~Zanfir, A.-I. Popa, A.~Zanfir, and C.~Sminchisescu, ``Human appearance transfer,'' in \emph{The IEEE Conference on Computer Vision and Pattern Recognition (CVPR)}, 2018.

\bibitem{neverova2018dense}
N.~Neverova, R.~A. Guler, and I.~Kokkinos, ``Dense pose transfer,'' in \emph{Proceedings of the European conference on computer vision (ECCV)}, 2018, pp. 123--138.

\bibitem{li2019dense}
Y.~Li, C.~Huang, and C.~C. Loy, ``Dense intrinsic appearance flow for human pose transfer,'' in \emph{Proceedings of the IEEE/CVF Conference on Computer Vision and Pattern Recognition}, 2019, pp. 3693--3702.

\bibitem{li2020pona}
K.~Li, J.~Zhang, Y.~Liu, Y.-K. Lai, and Q.~Dai, ``Pona: Pose-guided non-local attention for human pose transfer,'' \emph{IEEE Transactions on Image Processing}, vol.~29, pp. 9584--9599, 2020.

\bibitem{gafni2020wish}
O.~Gafni and L.~Wolf, ``Wish you were here: Context-aware human generation,'' in \emph{The IEEE Conference on Computer Vision and Pattern Recognition (CVPR)}, 2020.

\bibitem{ioffe2015batch}
S.~Ioffe and C.~Szegedy, ``{B}atch {N}ormalization: Accelerating deep network training by reducing internal covariate shift,'' in \emph{The International Conference on Machine Learning (ICML)}, 2015.

\bibitem{nair2010rectified}
V.~Nair and G.~E. Hinton, ``Rectified linear units improve restricted boltzmann machines,'' in \emph{The International Conference on Machine Learning (ICML)}, 2010.

\bibitem{gulrajani2017improved}
I.~Gulrajani, F.~Ahmed, M.~Arjovsky, V.~Dumoulin, and A.~Courville, ``Improved training of wasserstein gans,'' \emph{arXiv preprint arXiv:1704.00028}, 2017.

\bibitem{cao2017realtime}
Z.~Cao, T.~Simon, S.-E. Wei, and Y.~Sheikh, ``Realtime multi-person 2d pose estimation using part affinity fields,'' in \emph{The IEEE Conference on Computer Vision and Pattern Recognition (CVPR)}, 2017.

\bibitem{he2016deep}
K.~He, X.~Zhang, S.~Ren, and J.~Sun, ``Deep residual learning for image recognition,'' in \emph{The IEEE Conference on Computer Vision and Pattern Recognition (CVPR)}, 2016.

\bibitem{simonyan2015very}
K.~Simonyan and A.~Zisserman, ``Very deep convolutional networks for large-scale image recognition,'' in \emph{The International Conference on Learning Representations (ICLR)}, 2015.

\bibitem{li2017multiple}
J.~Li, J.~Zhao, Y.~Wei, C.~Lang, Y.~Li, T.~Sim, S.~Yan, and J.~Feng, ``Multiple-human parsing in the wild,'' \emph{arXiv preprint arXiv:1705.07206}, 2017.

\bibitem{liu2016deepfashion}
Z.~Liu, P.~Luo, S.~Qiu, X.~Wang, and X.~Tang, ``{D}eep{F}ashion: Powering robust clothes recognition and retrieval with rich annotations,'' in \emph{The IEEE Conference on Computer Vision and Pattern Recognition (CVPR)}, 2016.

\bibitem{kingma2015adam}
D.~P. Kingma and J.~Ba, ``Adam: A method for stochastic optimization,'' in \emph{The International Conference on Learning Representations (ICLR)}, 2015.

\bibitem{wang2004image}
Z.~Wang, A.~C. Bovik, H.~R. Sheikh, and E.~P. Simoncelli, ``Image quality assessment: From error visibility to structural similarity,'' \emph{IEEE Transactions on Image Processing (TIP)}, 2004.

\bibitem{salimans2016improved}
T.~Salimans, I.~J. Goodfellow, W.~Zaremba, V.~Cheung, A.~Radford, and X.~Chen, ``Improved techniques for training {GAN}s,'' in \emph{The Conference on Neural Information Processing Systems (NIPS)}, 2016.

\bibitem{liu2016ssd}
W.~Liu, D.~Anguelov, D.~Erhan, C.~Szegedy, S.~Reed, C.-Y. Fu, and A.~C. Berg, ``{SSD}: Single shot multibox detector,'' in \emph{The European Conference on Computer Vision (ECCV)}, 2016.

\bibitem{andriluka20142d}
M.~Andriluka, L.~Pishchulin, P.~Gehler, and B.~Schiele, ``{2D} human pose estimation: New benchmark and state of the art analysis,'' in \emph{The IEEE Conference on Computer Vision and Pattern Recognition (CVPR)}, 2014.

\bibitem{szegedy2015going}
C.~Szegedy, W.~Liu, Y.~Jia, P.~Sermanet, S.~Reed, D.~Anguelov, D.~Erhan, V.~Vanhoucke, and A.~Rabinovich, ``Going deeper with convolutions,'' in \emph{The IEEE Conference on Computer Vision and Pattern Recognition (CVPR)}, 2015.

\bibitem{kullback1951information}
S.~Kullback and R.~A. Leibler, ``On information and sufficiency,'' \emph{The Annals of Mathematical Statistics}, 1951.

\bibitem{zhang2018unreasonable}
R.~Zhang, P.~Isola, A.~A. Efros, E.~Shechtman, and O.~Wang, ``The unreasonable effectiveness of deep features as a perceptual metric,'' in \emph{The IEEE Conference on Computer Vision and Pattern Recognition (CVPR)}, 2018.

\bibitem{iandola2016squeezenet}
F.~N. Iandola, S.~Han, M.~W. Moskewicz, K.~Ashraf, W.~J. Dally, and K.~Keutzer, ``{S}queeze{N}et: {A}lex{N}et-level accuracy with 50x fewer parameters and <0.5mb model size,'' \emph{arXiv preprint arXiv:1602.07360}, 2016.

\end{thebibliography}

\end{document}